\pdfoutput=1

\documentclass[11pt]{article}

\usepackage[final]{emnlp2021}

\usepackage{times}
\usepackage{latexsym,amsmath,amssymb,amstext}
\usepackage{graphicx}
\usepackage{subfigure}
\usepackage{paralist}
\usepackage{multirow}
\usepackage[T1]{fontenc}

\usepackage[utf8]{inputenc}
\usepackage{arydshln}

\usepackage{microtype}
\usepackage{color}

\usepackage{amsmath,amsfonts,bm}
\usepackage{mathtools}

\def\eqref#1{equation~\ref{#1}}

\def\1{\bm{1}}

\def\rvv{{\mathbf{v}}}

\def\rvx{{\mathbf{x}}}

\def\rmM{{\mathbf{M}}}

\def\rmW{{\mathbf{W}}}
\def\rmX{{\mathbf{X}}}

\DeclareMathAlphabet{\mathsfit}{\encodingdefault}{\sfdefault}{m}{sl}
\SetMathAlphabet{\mathsfit}{bold}{\encodingdefault}{\sfdefault}{bx}{n}

\def\gD{{\mathcal{D}}}

\def\gY{{\mathcal{Y}}}

\def\sR{{\mathbb{R}}}

\newcommand{\softmax}{\mathrm{softmax}}

\def\softmax{{\mbox{softmax}}}
\def\FFN{{\mbox{FFN}}}
\def\AvgPool{{\mbox{AvgPool}}}
\def\MaxPool{{\mbox{MaxPool}}}

\title{Sequential Attention Module for Natural Language Processing}

\author{Mengyuan Zhou, Jian Ma, Haiqin Yang$^{\S}$, Lianxin Jiang, Yang Mo \\
         Ping An Life Insurance, Ltd. \\
  Shenzhen City, Guangdong Province, China \\
         {\small\{ZHOUMENGYUAN425,MAJIAN446,JIANGLIANXIN769,MOYANG853\}@pingan.com.cn}
         \\$^{\S}$ {\small the corresponding author, email: {hqyang@ieee.org}}
     }

\begin{document}
\maketitle
\begin{abstract}
Recently, large pre-trained neural language models have attained remarkable performance on many downstream natural language processing (NLP) applications via fine-tuning.  In this paper, we target at how to further improve the token representations on the language models.  We, therefore, propose a simple yet effective plug-and-play module, Sequential Attention Module (SAM), on the token embeddings learned from a pre-trained language model.  Our proposed SAM consists of two main attention modules deployed sequentially: Feature-wise Attention Module (FAM) and Token-wise Attention Module (TAM).  More specifically, FAM can effectively identify the importance of features at each dimension and promote the effect via dot-product on the original token embeddings for downstream NLP applications.  Meanwhile, TAM can further re-weight the features at the token-wise level.  Moreover, we propose an adaptive filter on FAM to prevent noise impact and increase information absorption.  Finally, we conduct extensive experiments to demonstrate the advantages and properties of our proposed SAM.   We first show how SAM plays a primary role in the champion solution of two subtasks of SemEval'21 Task 7.  After that, we apply SAM on sentiment analysis and three popular NLP tasks and demonstrate that SAM consistently outperforms the state-of-the-art baselines.  
\end{abstract} 

\section{Introduction}
In natural language processing (NLP), pretraining large neural language models on a plethora of unlabeled text have proven to be a successful strategy for transfer  learning~\cite{DBLP:conf/nips/YangDYCSL19,DBLP:conf/iclr/ClarkLLM20}.  An elite example, e.g., Bidirectional Encoder Representations from Transformers  (BERT)~\cite{DBLP:conf/naacl/DevlinCLT19}, or RoBERTa~\cite{DBLP:journals/corr/abs-1907-11692}, has become a standard building block for training task-specific NLP models via fine-tuning.

Transformer~\cite{DBLP:conf/nips/VaswaniSPUJGKP17} has become a backbone architecture in NLP to model the dependency of long-distance sequences via an attention mechanism based on a function that operates on queries, keys, and values.  The attention function maps a query and a set of key-value pairs to an output, where the output is a weighted sum of the values.  For each value, a weight is computed as a compatibility function of the query with the corresponding key.  The mechanism can be viewed as projecting the input sequences to multiple subspaces while concatenating the output of the scaled dot-product attention to the representation in each subspace afterward.  Information embedded in the sequence of tokens can then be effectively absorbed.  However, the existing attention mechanism mainly absorbs the information at the token-wise level and ignores the information at the feature-wise (or multi-dimensional) level~\cite{DBLP:conf/emnlp/SinhaDCR18,DBLP:conf/emnlp/ChenZ18,DBLP:conf/emnlp/Cao000L18}.  

In this paper, to further improve the information absorption, we propose a  novel and simple plug-and-play module, Sequential Attention Module (SAM), on the token embeddings of the pre-trained language model.  More specifically, our proposed SAM sequentially deploys two attention sub-modules, the feature-wise attention module (FAM) and the token-wise attention module (TAM).  FAM can effectively identify the importance of features at each dimension to resolve the polysemy issue and promote the effect via dot-product on the original token embeddings for downstream NLP applications.   Meanwhile, TAM can further re-weight the features at the token-wise level while expanding the search space of the attention parameters.  Hence, SAM can work as the procedure of sentence construction, building words from basic features and constructing sentences based on words.  Moreover, we propose an adaptive filter on FAM to prevent noise impact and increase information absorption.  After obtaining the re-weighted features via SAM, we can feed them to a standard classifier, e.g., softmax, to solve the downstream classification tasks.  

We highlight the contribution of our work in the following:
\begin{compactitem}
\item  We propose a simple yet effective plug-and-play attention module, Sequential Attention Module (SAM), which can absorb the information at both the feature-wise level and token-wise level. 

\item We further design an adaptive filter on the feature-wise attention module prevent noise impact and increase information absorption. 

\item We conduct extensive experiments to demonstrate the advantages and properties of our proposed SAM.  We first show how SAM plays a critical role in the champion solution of two subtasks of SemEval'21 Task 7.  The results on sentiment analysis and three popular NLP tasks show that SAM outperforms the state-of-the-art baselines significantly and prove the generalization power of SAM.  Our code is released and can publicly downloaded in~\url{https://www.dropbox.com/s/l0n39tvxjmq0k4l/SAM.zip?dl=0}. 


\end{compactitem}



\section{Related Work}
In the following, we review related work on pre-trained language models (PLM) and attention mechanisms on NLP applications.  

{\bf PLM.} Recently, in natural language processing, large neural pre-trained language models, such as BERT, ELECTRA~\cite{DBLP:conf/iclr/ClarkLLM20}, XLNet~\cite{DBLP:conf/nips/YangDYCSL19}, and open-GPT~\cite{DBLP:journals/crossroads/CohenG20}, have been developed to attain precise token (sub-words) representations, which are learned from a plethora of unlabeled text.  After that, fine-tuning is applied to update the pre-trained word representations for downstream tasks with a small number of task-specific parameters~\cite{DBLP:conf/naacl/DevlinCLT19,DBLP:journals/corr/abs-1907-11692}.  Due to the heavy overhead of the pre-trained language models, researchers have tried various ways to reduce the computation during inference.  For example, distillation has been applied to compress a large teacher model to a small student while maintaining the model performance~\cite{DBLP:journals/corr/abs-1910-01108,DBLP:conf/emnlp/JiaoYSJCL0L20,DBLP:conf/acl/SunYSLYZ20,DBLP:conf/acl/LiuZWZDJ20}.  Interaction between tokens has been delayed to reduce the computation in the self-attention mechanism~\cite{DBLP:conf/iclr/HumeauSLW20,DBLP:conf/sigir/KhattabZ20}.  However, little work has attempted to explore the potential of utilizing the information embedded in the well-trained token representations.  


Attention mechanism has been largely deployed in many NLP applications and exploited in different ways: (1) \textbf{weight calculation}: Basic operations, such as dot-product, general weighting, and concatenation with non-linear functions, have been attempted to compute the attention weight for neural machine translation~\cite{DBLP:journals/corr/BahdanauCB14,DBLP:conf/emnlp/LuongPM15}.  (2) \textbf{Information fusion}:  Several attention mechanisms, e.g., co-attention~\cite{DBLP:conf/iclr/XiongZS17}, gated-attention~\cite{DBLP:conf/acl/DhingraLYCS17}, and bidirectional attention~\cite{DBLP:conf/iclr/SeoKFH17}, have been attempted to enhance the performance of reading comprehension.  (3) {\bf Structure leverage}: Hierarchical  attention~\cite{DBLP:conf/ijcnlp/PappasP17} has been designed to absorb both the word-level and the sentence-level information for text classification.  {Two-stream self-attention mechanism} has been invented in XLNETfor computing the target-aware representations~\cite{DBLP:journals/corr/abs-1906-08237}.  Coupled multi-Layer attentions have been proposed to co-extract aspect and opinion terms~\cite{DBLP:conf/aaai/WangPDX17}.  A self-attention mechanism has been applied to obtain more precise sentence representations~\cite{DBLP:conf/iclr/LinFSYXZB17}.  Multi-hop attentions have been applied to establish the interaction between the
question and the context~\cite{DBLP:conf/acl/GongB18}.  Directional Self-Attention Network (DiSAN) has been deployed on RNN/CNN to include both directional and multi-dimensional information~\cite{DBLP:conf/aaai/ShenZLJPZ18}.   However, the existing attention mechanisms have not been investigated to improve the token representations on the pre-trained language models.

\section{Our Proposal}

In the following, we first define the problem.  Next, we present our proposed play-and-plug module, Sequential Attention Module (SAM), which consists of two sequential attention modules, Feature-wise Attention Module (FAM) and Token-wise Attention Module (TAM).  

\begin{figure*}[t]
\centering
\includegraphics[width=0.9\linewidth]{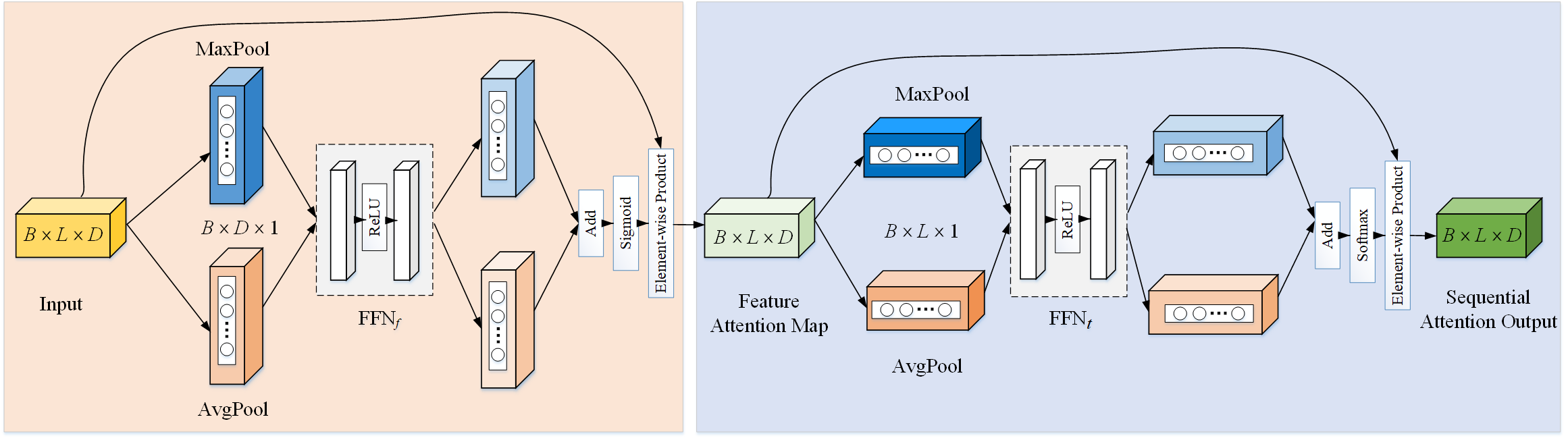}
\caption{The overview of sequential attention module. \label{fig:SAM}}
\end{figure*}

\subsection{Notations and Problem Definition}
{\bf Notations.} To make the notation consistent, we let lowercase denote a vector, a bold lowercase denote a sequence of vectors (stored as a matrix), and a bold uppercase denote a matrix or a tensor.  To avoid notation redundancy, we define a fully connected feed-forward network, $\FFN$, which consists of two linear transformations with a ReLU activation in between as in Transformer~\cite{DBLP:conf/nips/VaswaniSPUJGKP17}:   
\begin{equation}\label{eq:FFN}
\FFN(\rvx) = \max(0, \rvx\rmW_1+b_1)\rmW_2+b_2,
\end{equation}
where $\rmW_1$ and $\rmW_2$ are the corresponding weight parameters,  {$b_1$ and $b_2$ and the corresponding bias parameters.}

{\bf Setting.} In an NLP task, we are given a set of labeled sentences, $\gD=\{(\rvv_i, y_i)\}_{i=1}^N$, where $N$ is the number of training data, $\rvv_i$ denotes the $i$-th sentence, which consists of a sequence of $|\rvv_i|$ tokens (subwords or characters), or $\rvv_i=v_{i1}\,v_{i2}\,\ldots\,v_{i|\rvv_i|}$.  The $i$-th response for $\rvv_i$ is $y_i\in\gY$, where $\gY=\{-1, +1\}$ for a binary classification task and $\gY=\{1, 2, \ldots, K\}$ for a $K$-class classification task.  $v_{ij}$ ($j=1, \ldots, |\rvv_i|$) could be a one-hot vector whose dimension length equals the number of distinct tokens in the dictionary.  The sequence of $\rvv$ can be represented by $\rvx = g(\rvv)\in\sR^{D\times |\rvx|}$, where $D$ is the embedding size and $g$ yields the output of the token representation by 
\begin{compactitem}
\item a pre-trained language model, e.g., BERT or  RoBERTa; 
\item the output of the pre-trained language model with additional operations, e.g., a Bidirectional Long Short-Term Memory (BiLSTM)~\cite{DBLP:conf/acl/ZhouSTQLHX16}.  This allows to capture the high-level semantic information of tokens.
\end{compactitem}

{\bf Problem.} Suppose the mini-batch size of the data is $B$, the maximum length of the sequence to $L$, we can represent the data by a tensor $\rmX\in\sR^{B\times L\times D}$ as shown in the input (the leftmost block) of Fig.~\ref{fig:SAM}.  Our {goal} is to further utilize the information embedded in the tensor to improve the performance on the downstream tasks.  


\subsection{Feature-wise Attention Module (FAM)}
First, SAM infers a {\bf feature-wise attention map} to identify the feature importance at each dimension:  
\begin{equation}\label{eq:FAM}
\begin{aligned}
    {{\rmM}_{f}}(\rmX) = \sigma(&\FFN_f(\MaxPool_f(\rmX))\\
&+\FFN_f(\AvgPool_f(\rmX))),
\end{aligned}
\end{equation}
where $\MaxPool_f$ and $\AvgPool_f$ are the feature-wise maximum pooling and average pooling on the input $\rmX$, respectively.  Hence, by the pooling operations, we obtain two tensors with the size of $B\times D\times 1$ and feed them to a shared feature-wise fully connected feed-forward network, $\FFN_f$, as defined in Eq.~(\ref{eq:FFN}), to identify the feature importance.  

\paragraph{Adaptively filtered feature-wise attention map (AF-FAM).}  To prevent noise effect, we propose an {\bf adaptively filtered feature-wise attention map} and derive the output as follows: 
\begin{align}\label{eq:AF-FAM}
{{\rmM}^{\delta}_{f}}(\rmX) &= \max(0, {{\rmM}_{f}}(\rmX)-\delta), \\
  {\rmX}'&={{\rmM}^{\delta}_{f}}(\rmX)\otimes \rmX, 
\end{align}
where $\delta\in[0, 1]$ is a constant to mask the corresponding feature.  When $\delta=0$, the output of ${\rmX}'$ is re-weighted by the feature-wise attention map, which is determined originally by Eq.~(\ref{eq:FAM}).  When $\delta=1$, since the output of Eq.~(\ref{eq:FAM}) is in the range of 0 and 1, AF-FAM will mask out all of the features and prevent the information flowing to the next step.  If $\delta$ is a small constant, it allows AF-FAM to filter out the uncertain features, which can prevent the noise effect.  $\delta$ can also play an effective way to explain the importance of the hidden features~\cite{DBLP:conf/ijcnlp/DanilevskyQAKKS20} or can be automatically tuned by adversarial training~\cite{DBLP:journals/corr/abs-2004-08994,DBLP:journals/tist/ZhangSAL20}.  We leave these investigation as future work. 

\subsection{Token-wise Attention Module (TAM)} 
Second, SAM defines a {\bf token-wise attention map} to identify the importance of $\rmX'$ at the token-wise level: 
\begin{equation}
\begin{aligned}
    {{\rmM}_{t}}(\rmX') = &\softmax(\FFN_t(\MaxPool_t(\rmX'))\\
&+\FFN_t(\AvgPool_t(\rmX'))),
\end{aligned}
\end{equation}
where $\MaxPool_t$ and $\AvgPool_t$ are similar to $\MaxPool_f$ and $\AvgPool_f$, but conduct the maximum pooling and average pooling at the token-wise level and yield two other tensors with the size of $B\times L\times 1$, respectively.  $\FFN_t$ is similarly defined in Eq.~(\ref{eq:FFN}) at the token-wise level.  The operation $\softmax$ is applied to normalize the weight. 

The final output is then computed by 
\begin{equation}\label{eq:TWAM_DP}
 {\rmX}''={{\rmM}_{t}}({\rmX}')\otimes {\rmX}'.
\end{equation}
Here, we do not conduct information filtering on the $\softmax$ operation in ${{\rmM}_{t}}(\rmX')$ for fidelity.  The exploration of the effect filtering $\softmax$ and other transformations, e.g., pooling, can be further investigated in the future work.  

\subsection{Integration and Classification}
As illustrated in Fig.~\ref{fig:SAM}, SAM sequentially integrates two attention modules, FAM and TAM.  Obviously, these two modules can be swapped.  Our experimental evaluation shows that FAM follows by TAM can attain better performance than the alternate one.  We provide more discussions in Sec.~\ref{sec:exp}.    

Hence, after we obtain the features via SAM in Eq.~(\ref{eq:TWAM_DP}), we can integrate them into a classification model, e.g., softmax, via minimizing the cross-entropy loss to train the model parameters for downstream classification tasks.  We leave the exploration of applying SAM on other tasks, e.g., regression tasks, as future work.


\section{Experiments}\label{sec:exp}


In the following, we first present the experimental setup and the evaluation on two classification subtasks on SemEval-2021 Task 7~\cite{meaney2021hahackathon} because our proposed SAM is the main module of the champion solution~\footnote{\url{http://smash.inf.ed.ac.uk/tasks_results/hahackathon_results.html}}.  Next, we test SAM on sentiment analysis and three other popular NLP tasks to verify the generalization ability.  After that, we conduct ablation study on SAM to test the effect of different modules and combinations.  We also present sensitivity analysis on $\delta$ and deliver case study to exhibit the properties of SAM.    



\begin{table}[!htb]
\centering
\begin{tabular}{p{1.5cm}lll}
\hline 
    \textbf{Dataset}& \textbf{Train}& \textbf{Dev}&  \textbf{Test}\\
\hline
    {\bf Humor.} &8,000&1,000&1,000\\
    (binary) & 69/25.7/3& 79/23.9/3& 70/28.1/4 \\\hdashline
    {\bf Contro.} & 4,935& 1,000& 1,000\\
    (binary) & 69/25.7/3 & 79/23.9/3 & 70/28.1/4\\\hdashline
    {\bf SST}& 8,544& 1,101& 2,210\\
    (5-class) & 52/21.4/2 & 50/20.6/2 & 57/20.7/2\\\hdashline
    {\bf CR}& 3,016& & 755\\
    (binary)& 106/20.7/1 &  & 112 /20.5/1 \\\hdashline
    {\bf SUBJ}& 8,000& & 2,000\\
     (binary)& 58/25.0/10 &  & 59/25.2/10 \\\hdashline
    {\bf TREC}& 5,452& & 500\\
     (6-class)& 29/10.2/4 &  & 22/7.8/4\\\hline
\end{tabular}
\caption{Data statistics: each dataset consists of two rows; the first row records the number of samples in the splitting sets while the second row records the maximum, average, and minimum number of sub-words of each sentence in the corresponding sets.  To save the space, we record the number of classes with the round bracket under the dataset name.  For example, binary implies 2-class. \label{tab:data}}
\end{table}

\subsection{Experimental Setup}
{RoBERTa}is adopted as the backbone language model because it is competitive and has been applied to attain champion solutions for many NLP competitions.  The AdamW optimizer~\cite{DBLP:journals/corr/abs-1711-05101} is adopted with a batch size of 32.  The lookahead mechanism~\cite{DBLP:conf/nips/ZhangLBH19} is applied to help AdamW to accelerate the training.  The weight decaying rate is 1e-2.  The BiLSTM module consists of 2 layers with the size of the hidden units being 256.  All models are implemented with Pytorch and run on a single NVIDIA GeForce RTX 3090 graphic card.  

\subsection{SemEval-2021: HaHackathon}\label{sec:HaHackathon}
The SemEval-2021 Task 7~\cite{meaney2021hahackathon} consists of four subtasks to detect or rate humor and offense in different aspects: (1) Task 1a ({\bf Humor.}) is to predict if the text would be considered humorous for an average user.  This is a binary classification task.  (2) Task 1b ({\bf Contro.}) is to predict how humorous it is for an average user when the text is classed as humorous.  This is a regression task to predict a value in the range of 0 and 5. (3) Task 1c is to predict if the humor rating would be considered controversial, i.e. the variance of the rating between annotators is higher than the median, when the text is classed as humorous.  It is also a binary classification task.  (4) Task 2 aims to predict how offensive a text would be for an average user.  It is a regression task to predict a value in the range of 0 and 5.  Among these four tasks, we report the evaluation results on the binary classification sub-tasks because we attain the champion on them.  The statistics of the tasks are listed in the first four rows of Table~\ref{tab:data}.  A main challenge of the tasks is that humor detection is highly subjective.

In the competition, we evaluation the following methods: (1) {\bf RoBERTa}: RoBERTa is a competitive backbone language model; (2) {\bf RB}: BiLSTM is deployed on top of the RoBERTa feature extractor; (3) {\bf RBA}: Attention-based BiLSTM~\cite{DBLP:conf/acl/ZhouSTQLHX16} is applied on top of the RoBERTa feature extractor; (4) {\bf SAM}: our proposed SAM is built on {\bf RB}.  Without specifying, $\delta$ is set to 0.0 for simplicity.  All compared methods are fine-tuned by optimizing the initial learning rate from 1.0e-5 to 1.4e-5 with the step size of 0.1e-5 in the {\bf Humor.} dataset and from 1.0e-5 to 1.3e-5 with the step size of 0.1e-5 in the {\bf Contro.} dataset, respectively, while the dropout rate is tried from 0.2 to 0.4 with the step size of 0.1 in the {\bf Humor.} dataset and from 0.1 to 0.3 with the step size of 0.1 in the {\bf Contro.} dataset for trial and error. The maximum number of epochs is set to 50.  F1-score is the evaluation metric to determine the performance in the competition.

\begin{table}[!htb]
\centering
\begin{tabular}{lcccc}
\hline 
    \multirow{2}{*}{\textbf{Methods}} & \multicolumn{2}{c}{\bf Humor. } &  \multicolumn{2}{c}{\bf Contro. } \\\cline{2-5}
    &{\textbf{F1}}  &{\textbf{T(s)}}&{\textbf{F1}}  &{\textbf{T(s)}}\\
\hline
    {\bf RoBERTa} & {95.42} & {83} & {59.29} & {51}\\
    {\bf RB} & {95.48} & {92} & {59.06} & {57}\\
    {\bf RBA} & {95.68} & {92} & {60.33} & {57} \\
    {\bf SAM} & {{\bf 96.41}}  & {92}& {{\bf 62.89}}  & {57}\\
\hline
\end{tabular}
\caption{Experimental results on SemEval-2021: HaHackathon.  The F1-scores ({\bf F1}) and the time trained at each epoch, denoted by {\bf T(s)} in seconds, are averaged on 5-fold cross-validation. \label{tab:SemEval21-HaHackathon}}
\end{table}
Table~\ref{tab:SemEval21-HaHackathon} reports the average F1-scores of 5-fold cross-validation on two subtasks, respectively.  The results show that 
\begin{compactitem}
\item RB and RBA consistently increase the performance gradually.  This indicates the token representations extracted by RoBERTa can be further utilized. 
\item SAM obtains significant better improvement ($p<0.01$ in paired $t$-test) than the compared methods.  It is noted that the reported F1-scores are slightly lower than those in the counter parts, 98.54 at Task 1a and 63.02 at Task 1c, of the competition leaderboard.  The reason lies that for the champion solution, we have conducted additional tricks, e.g., model ensemble\cite{DBLP:journals/asc/ForoughM21}
, label smoothing\cite{DBLP:conf/nips/MullerKH19}, and pseudo labeling\cite{DBLP:journals/corr/abs-2103-03335}.  However, we argue that SAM is the main module to attain the exceptional performance. 
\item 
By examining the recorded time cost of each epoch, we observe that SAM consumes the same time as RB and RBA, which implies the efficiency of SAM.
\end{compactitem}


\subsection{Sentiment Analysis}\label{sec:SA}
Emotion is a high-level intelligent behavior of  humans~\cite{DBLP:conf/wassa/Strapparava16,DBLP:conf/emnlp/BuonoSBGTM17,DBLP:conf/nlpcc/WuJ19}.  It aims at developing models to understand the emotional tendency of texts at the semantic level.  The difficulty of sentiment analysis lies in identifying the polysemous meaning of a word, which needs to absorb both the token-wise information and distinguish the ambiguity in different contexts.  

Stanford Sentiment Treebank~\cite{DBLP:conf/emnlp/SocherPWCMNP13} is a standard sentiment analysis dataset released by the NLP group of Stanford University.  It mainly used for sentiment classification, in which each node of the sentence analysis tree has fine-grained sentiment annotations.  The standard train/dev/test sets are split into 8,544/1,101/2,210 samples and the statistics of the dataset is recorded in the fifth and sixth rows of Table~\ref{tab:data}. 

\begin{table}[ht]
\centering
\begin{tabular}{lcc}
\hline 
    \textbf{Methods} &{\textbf{Acc.}}  &{\textbf{T(s)}}\\
\hline
    {\bf RoBERTa} & {59.10} & {97} \\
    {\bf RB} & {59.30} & {112} \\
    {\bf RBA} & {59.40} & {112} \\
    {\bf DiSAN} & {59.20}  & {211}\\
    {\bf SAM } & {{\bf 59.70}}  & {112}\\
\hline
\end{tabular}
\caption{Experimental results on the Stanford Sentiment Tree-bank (SST).   The accuracy ({Acc.}) is averaged of five runs on the test set and the time trained at each epoch, denoted by {\bf T(s)} in seconds, is the average of five runs.\label{tab:rs_SST} 
}
\end{table}
Other than the four compared methods depicted in Sec.~\ref{sec:HaHackathon}, we also conduct experiments on Directional Self-Attention Network (DiSAN)~\cite{DBLP:conf/aaai/ShenZLJPZ18}, a light-weight network showing strong performance on RNNs/CNNs via includeing multi-dimensional (or feature-wise) information.  We place DiSAN on top of RoBERTa.  All compared methods are fine-tuned by optimizing the initial learning rate from 1.0e-3 to 1.5e-3 with the step size of 1.0e-3 and the dropout rate is tried from 0.2 to 0.5 with the step size of 0.1 for trial and error. The maximum number of epochs is set to 50.  Following, we evaluate the accuracy on the test set and report the time in seconds in Table~\ref{tab:rs_SST}.  The results show that 
\begin{compactitem}
\item The higher accuracy scores in RB, RBA, and DiSAN than RoBERTa imply the effectiveness of the compared methods in further absorbing token embeddings from RoBERTa. 
\item SAM can further improve the performance and beats RBA, the best baseline, at least 0.5\% improvement. 
\item Similarly, the time cost of each epoch for SAM is the same time as that in RB and RBA.  On the contrary, DiSAN consumes nearly double time but yields worse than SAM.
\end{compactitem}


\begin{table*}[!htb]
\centering
\begin{tabular}{lc@{~}cc@{~}cc@{~}cc@{~}cc@{~}cc@{~}c}
\hline 
    \multirow{2}{*}{\textbf{Settings}} & \multicolumn{2}{c}{\bf Humor. } &  \multicolumn{2}{c}{\bf Contro. } & \multicolumn{2}{c}{\bf SST } & \multicolumn{2}{c}{\bf CR} & \multicolumn{2}{c}{\bf SUBJ }& \multicolumn{2}{c}{\bf TREC} \\\cline{2-13}
    &{\textbf{F1}}&{\textbf{T(s)}} &{\textbf{F1}}&{\textbf{T(s)}} 
    &{\textbf{Acc.}}&{\textbf{T(s)}} &{\textbf{Acc.}}&{\textbf{T(s)}} 
    &{\textbf{Acc.}}&{\textbf{T(s)}} &{\textbf{Acc.}}&{\textbf{T(s)}} \\
\hline
    {\bf RoBERTa} & {95.42} & {83} & {59.29} & {51} & {59.10} & {97} & {87.56} & {14} & {94.09} & {40} & {91.90} & {8}\\\hdashline

    {\bf SAM ($-$FAM)} & {95.70} & {91} & {60.50} & {58} & {59.30} & {105} & {89.77} & {16} & {95.69} & {41} & {92.60} & {10}\\
     {\bf SAM ($-$TAM)} & {95.50} & {91} & {60.65} & {58} &{59.10} & {105} &
     {89.02} & {16} &{94.99} & {41} &{92.40} & {10} \\
     
     {\bf SAM (TAM+FAM)} & {95.98} & {92} & {61.03} & {57} &
     {58.45} & {112} &{89.91} & {16} &
     {{95.83}} & {43} &{93.20} & {11} \\

     {\bf SAM }($\delta=0.1$) & {96.12} & {93} & {60.86} & {56} &
     {59.40} & {114} &{89.97} & {17} &
     {\bf 95.84} & {45} &{93.20} & {11} \\
     
    {\bf SAM } & {{\bf 96.41}}  & {92}& {\bf 61.22}  & {57}
    & {{\bf 59.70}}  & {112}& {{\bf 90.03}}  & {16}
    & {95.78}  & {43}& {\bf 93.40}  & {11}\\
\hline
\end{tabular}
\caption{Ablation study on SAM under different settings.\label{tab:SAM_AS}} 
\end{table*}
\subsection{Experiments on Other NLP Tasks}\label{sec:NLP}
To verify the generalization of our proposed SAM, we conduct experiments on three popular NLP tasks: (1) \textbf{CR}~\cite{DBLP:conf/kdd/HuL04}: Customer review of various products, e.g., cameras, which is to predict whether the review is positive or negative; (2) \textbf{SUBJ}~\cite{DBLP:conf/acl/PangL04}: Subjectivity dataset whose labels indicate whether each sentence is subjective or objective; (3) \textbf{TREC}~\cite{DBLP:conf/coling/LiR02}: The TREC dataset includes six different types of questions. There are 5452 labeled questions in the training set and 500 questions in the test set. The last six rows of Table~\ref{tab:data} record the corresponding data statistics.  It is shown that we also include multi-class classification in the evaluation.

\begin{table}[htb]
\centering
\begin{tabular}{@{~}l@{~}c@{~}c@{~}c@{~}c@{~}c@{~}c}
\hline 
    \multirow{2}{*}{\textbf{Methods}} & \multicolumn{2}{c}{\bf CR } &  \multicolumn{2}{c}{\bf SUBJ } & \multicolumn{2}{c}{\bf TREC }\\\cline{2-7}
    &{\textbf{Acc.}}  &{\textbf{T(s)}} & {\textbf{Acc.}}  & {\textbf{T(s)}} & {\textbf{Acc.}}  &{\textbf{T(s)}} \\
\hline
    {\bf RoBERTa} & {87.56} & {14} & {94.09} & {40} & {91.9} & {8}\\
    {\bf RB} & {88.14} & {16} & {95.39} & {43} & {92.4} & {11}\\
    {\bf RBA} & {89.52} & {16} & {95.05} & {43} & {92.6} & {11} \\
    {\bf DiSAN} & {87.60} & {33} & {94.11} & {84} & {91.9} & {23}\\
    {\bf SAM } & {\bf 90.03} & {16}& {\bf 95.78} & {43} & {\bf93.4} & {11}\\
\hline
\end{tabular}
\caption{Experimental results on three popular NLP tasks: CR, SUBJ, and TREC.  The accuracy ({Acc.}) and the time trained at each epoch, denoted by {\bf T(s)} in seconds, are averaged on 5-fold cross-validation. \label{tab:rs_NLP}}
\end{table}
Here, we aim to test the effect of different attention mechanisms.  Hence, we apply RoBERTa as a feature extractor, i.e., freezing the RoBERTa parameters without fine-tuning.  All compared methods are fine-tuned by optimizing the initial learning rate from 1.25e-4 to 1.75e-4 with the step size of 0.05e-4 in the {CR} dataset, from 1.0e-4 to 1.5e-4 with the step size of 0.1e-4 in the SUBJ. dataset, and 1.0e-4 to 1.5e-4 with the step size of 0.1e-4 in the {SUBJ.} dataset, respectively, while the dropout rate is tried from 0.2 to 0.5 with the step size of 0.1 in the {CR} dataset and from 0.2 to 0.4 with the step size of 0.1 in both the {SUBJ} dataset and the TREC dataset for trial and error. The maximum number of epochs is set to 50.

Table~\ref{tab:rs_NLP} reports the average accuracy ({\bf Acc.}) and the recorded time cost of 5-fold cross-validation and shows that 
\begin{compactitem}
\item RB, RBA, DiSAN can further improve the classification performance over RoBERTa accordingly. 
\item SAM beats the best baselines on the corresponding datasets and attains outperforms the baselines with at least 0.4\% improvement ($p<0.01$ in paired $t$-test). 
\item The time cost by SAM is the same as RB and RBA while DiSAN costs double of the time.  
\end{compactitem}

\begin{figure*}[t]
\centering
\begin{tabular}{l@{~}l}
\raisebox{.35em}{\bf RBA} & \includegraphics[width=.57\linewidth]{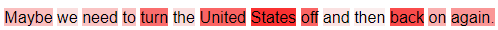} \\
\raisebox{.35em}{\bf SAM} & \includegraphics[width=.57\linewidth]{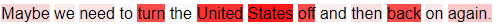} \\
\raisebox{.35em}{\bf RBA} & \includegraphics[width=.95\linewidth]{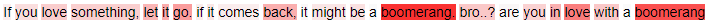} \\
\raisebox{.35em}{\bf SAM} & \includegraphics[width=.95\linewidth]{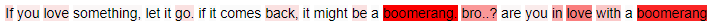} \\
\raisebox{.35em}{\bf RBA} & \includegraphics[width=.55\linewidth]{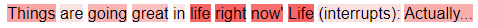} \\
\raisebox{.35em}{\bf SAM} & \includegraphics[width=.55\linewidth]{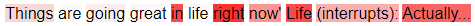} \\

\raisebox{.35em}{\bf RBA} & \includegraphics[width=.9\linewidth]{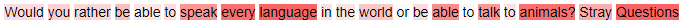} \\
\raisebox{.35em}{\bf SAM} & \includegraphics[width=.9\linewidth]{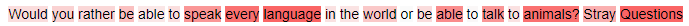} \\

\raisebox{.35em}{\bf RBA} & \includegraphics[width=.95\linewidth]{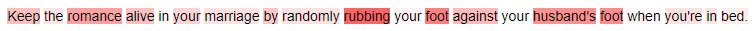} \\
\raisebox{.35em}{\bf SAM} & \includegraphics[width=.95\linewidth]{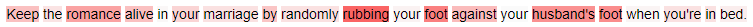} \\
\end{tabular}
\caption{Attention heatmap of {\bf RBA} and {\bf SAM} on five cases for Task 1a of SemEval-2021: HaHackathon.  Detailed description can be referred to the main text. \label{fig:attention_heatmap}
}
\end{figure*}
\subsection{Ablation Study}
We conduct detailed ablation study on SAM to thoroughly investigate its properties.  The evaluation settings include: (1) {\bf SAM ($-$FAM)}: we implement SAM by removing FAM to see its effect; (2) {\bf SAM ($-$TAM)}: we implement SAM by removing TAM to see its effect; (2) {\bf SAM (TAM+FAM)}: we implement SAM by swapping the order of FAM and TAM to see the effect of ; (3) {\bf SAM ($\delta=0.1$)}: we implement SAM by setting $\delta=0.1$; (4) {\bf SAM}: we implement SAM by setting $\delta=0.0$.  The implementation on the corresponding datasets follows the same setup as that in Sec.~\ref{sec:HaHackathon}, Sec.~\ref{sec:SA}, and Sec.~\ref{sec:NLP}] accordingly. 

Table~\ref{tab:SAM_AS} reports the performance on all six tasks and shows that
\begin{compactitem}
\item By comparing the results on {\bf SAM ($-$FAM)} and {\bf SAM ($-$TAM)}, we can see that SAM obtains worse performance than the complete one.  The performance on {\bf SAM ($-$FAM)} is slightly better than that on {\bf SAM ($-$TAM)}.  The observation shows that retaining the token-wise information will help the classification tasks.
\item Generally, swapping the order of FAM and TAM will decrease the performance slightly.  We conjecture that a sentence is composed by several tokens, where a tokens are composed by some features in the embeddings.  Hence, the token-wise information would be more helpful to determine the classification performance.  By swapping the attention order, we obtain the feature-wise information, which is weak for classification.

\item By varying $\delta$, we can further improve the model performance.  For example, when $\delta$ is set to 0.1, we can obtain the best performance on the SUBJ and TREC datasets.  More detailed investigation will be provided in Sec.~\ref{sec:SAD}. 
\end{compactitem}

\subsection{Sensitivity Analysis on $\delta$} \label{sec:SAD}
\begin{figure}[t]
\centering
\includegraphics[width=0.9\linewidth]{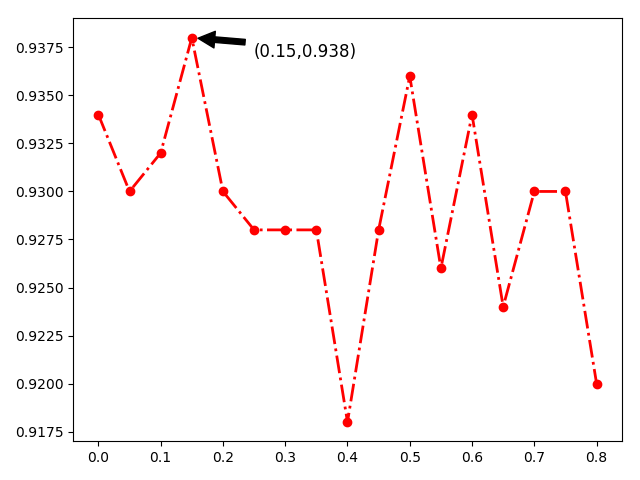}
\caption{Sensitivity analysis on $\delta$.\label{fig:Sensitivity analysis}}
\end{figure}

The parameter $\delta$ in Eq.~(\ref{eq:AF-FAM}) plays a key role to prevent noise effect.  When $\delta$ increases, FAM will try to mask the information on the corresponding dimension.  We test the effect of $\delta$ by varying it from 0.0 to 0.8 with the step size of 0.05.  Figure~\ref{fig:Sensitivity analysis} shows the results on the TREC dataset.  Other datasets follow similar characteristics.  The experimental results show that: (1) The best performance of SAM on the TREC dataset is attained when $\delta=0.15$.  The accuracy of 93.8 is higher than 93.4, the one obtained by the default SAM reported in Table~\ref{tab:rs_NLP}.   (2) When $\delta$ becomes larger, the performance of SAM usually decreases accordingly.  

\subsection{Case Study}
To gain a closer view of what information can be captured by SAM, we visualize the attention heatmap with respect to the classification model in Fig.~\ref{fig:attention_heatmap}, where dark color implies higher weight.  In particular, we compare our SAM with RBA, the best attention-based baseline, and show five sentences (the ground-truth for the first three sentences is $+1$, i.e., humorous, while the ground-truth for the last two sentences is $-1$, i.e., non-humorous).  It is noted that among these five cases, our SAM correctly predicts the labels while RBA mis-classifies them.  By observing the color intensity shown in Fig.~\ref{fig:attention_heatmap}, we can see that our SAM can further capture the role of important information to suppress non-important information, thereby enhancing the final effect.







\section{Conclusion}
In this paper, we propose a simple yet effective plug-and-play attention module, i.e., SAM, to further improve the token representations learned from pre-trained language models.  SAM follows a natural setting of language representation, from fine-grained features to high-level tokens.  That is, it first applies FAM to promote the feature importance.  After that, it applies TAM to enhance the search space of the attention mechanism.  Moreover, an adaptive filter is placed on FAM to prevent noise effect.  Our experimental results show that the proposed SAM is the key module to attain the champion solution in Task 1a and Task 1c of SemEval'21 Task 7.  Meanwhile, SAM exhibits its generalization power and consistently beats the state-of-the-art baselines on several popular NLP tasks, including sentiment analysis, customer review, subjective detection, question-answering selection.  Our ablation study shows that (1) the performance under the order of FAM following TAM is better than that under the alternate one, which again confirms the sentence construction procedure.  (2) the threshold parameter in FAM can be further tuned to improve the model performance.    

In the future, several significant problems are worthy of investigation on SAM: (1) How to automatically determine $\delta$ through adversarial training while enhancing the model explainability? (2) What are the potential ways to increase the effectiveness of information absorption in TAM? (3) What is the performance of SAM on other tasks, e.g., the regression task.   


\bibliography{custom}
\bibliographystyle{aclnatbib}

\end{document}